\documentclass[letterpaper, 10 pt, conference]{ieeeconf}  

\IEEEoverridecommandlockouts                         
\usepackage{graphicx}
\usepackage{amsmath}
\usepackage{amssymb}
\usepackage{booktabs}
\usepackage{marvosym}
\usepackage{multirow}
\usepackage{colortbl}
\usepackage{algorithm}
\usepackage{xcolor}
\usepackage{algorithm}
\usepackage{xcolor}
\usepackage{algpseudocode}
\usepackage{siunitx}
\usepackage{tabularx}
\usepackage[hang,flushmargin]{footmisc}

\usepackage{tikz}

\definecolor{cvprblue}{rgb}{0.21,0.49,0.74}
\definecolor{darkred}{RGB}{200, 0, 0}
\definecolor{darkgreen}{RGB}{0, 100, 0}

\usepackage[pagebackref,breaklinks,colorlinks,allcolors=cvprblue]{hyperref}

\title{\LARGE \bf ToSA: Token Merging with Spatial Awareness}

\author{Hsiang-Wei Huang, Wenhao Chai, Kuang-Ming Chen, Cheng-Yen Yang, and Jenq-Neng Hwang
\thanks{All the authors are from Electrical and Computer Engineering Department, University of Washington. Email: 
        {\tt\small \{hwhuang, wchai, kmchen, cycyang, hwang\}@uw.edu}}
}

\begin{document}

\maketitle

\begin{abstract}
Token merging has emerged as an effective strategy to accelerate Vision Transformers~(ViT) by reducing computational costs. However, existing methods primarily rely on the visual token's feature similarity for token merging, overlooking the potential of integrating spatial information, which can serve as a reliable criterion for token merging in the early layers of ViT, where the visual tokens only possess weak visual information. In this paper, we propose ToSA, a novel token merging method that combines both semantic and spatial awareness to guide the token merging process. ToSA leverages the depth image as input to generate pseudo spatial tokens, which serve as auxiliary spatial information for the visual token merging process. With the introduced spatial awareness, ToSA achieves a more informed merging strategy that better preserves critical scene structure. Experimental results demonstrate that ToSA outperforms previous token merging methods across multiple benchmarks on visual and embodied question answering while largely reducing the runtime of the ViT, making it an efficient solution for ViT acceleration. The code will be available at: \url{https://github.com/hsiangwei0903/ToSA}.
\end{abstract}

\section{Introduction}
Most of the recent vision foundation models~\cite{dinov2,clip,siglip} adopt Vision Transformer~(ViT)~\cite{vit} as the backbone, achieving advanced performance on various perception tasks such as classification, detection, and segmentation. On the other hand, these vision foundation models also play an important role in the field of generative AI, especially serving as the visual encoder of Vision Language Model~(VLM)~\cite{llava, onevision}. Although achieving great success, the attention mechanism of the ViT introduces heavy computational overhead and limits its further applications in real-world scenarios such as robotics, and autonomous driving, where high throughput and low computational cost are preferred.

Many previous works~\cite{efficientvit,dynamicvit,AViT} have explored more efficient ViT architecture to accelerate the runtime of ViT training and inference. These new architectures introduced extra learnable parameters or pooling layers that can reduce the number of visual tokens, therefore lowering the computational cost and runtime. Despite all these efforts, these methods require extra training due to their newly introduced model parameters, which limited their practicability compared with recent plug-and-play token reduction methods.

\begin{figure}[t]
    \centering
    \includegraphics[width=0.98\linewidth]{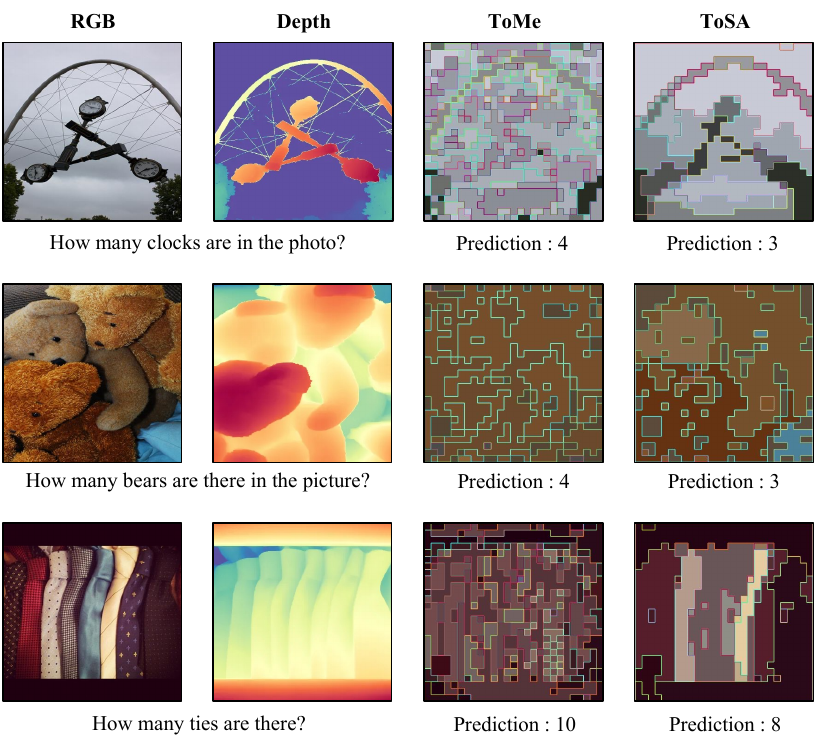}
    \caption{A merging comparison between ToMe and ToSA. By leveraging depth input, ToSA utilizes spatial awareness in the token merging process and leads to more spatially coherent merging results, helping the models to answer the question correctly. Both merging results retain 16 visual tokens for better visual comparison. Merged token is denoted by the same patch and inner edge color.}
    \label{fig:teaser}
\end{figure}

Recently, training-free token reduction methods have been introduced to improve the efficiency of VLM. Frame sampling~\cite{cheng2025tempura} or token pooling~\cite{tokenpooling,dtc} demonstrate effectiveness in reducing computational cost and the number of visual tokens in the ViT. Recent work such as ToMe~\cite{bolya2022tome} serves as a popular training-free method that largely reduces the computational cost of ViT, by introducing Bipartite Soft
Matching~(BSM), ToMe merges the visual token progressively in each layer of ViT based on the visual token features' similarity, achieving superior performance on image classification tasks compared with previous efficient ViT designs that require further training. Although these methods serve as an effective way to balance the trade-off between ViT efficiency and performance, they heavily rely on the visual features for the token merging and clustering process. Previous exploration work demonstrates~\cite{vitlearn} that the visual features in the early layer mostly capture low-level information such as edges and textures. For this reason, we argue the low-level features in the early layers are sub-optimal criteria for token merging, as the visual tokens from different objects can also demonstrate similarity in low-level visual features. In tasks such as embodied question answering or spatial question answering, preserving local information is also important, especially when facing questions related to counting, object existence...etc. For this reason, a token merging method that can preserve important local information for more fine-grained question answering task, and also enhance the ViT efficiency in VLM is needed.

In this work, we present \textbf{To}ken merging with \textbf{S}patial \textbf{A}wareness~(ToSA), aim to explore the potential of using the depth map from the RGB-D input as an auxiliary spatial awareness for training-free token merging. We compared ToSA with semantic-based token merging method ToMe~\cite{bolya2022tome} on various image and video question answering benchmarks including SpatialBench~\cite{spatialbot}, VQAv2-Counting~\cite{vqa}, GQA~\cite{gqa}, and OpenEQA~\cite{openeqa}. Experiment results show that state-of-the-art VLM incorporated with ToSA can demonstrate better performance compared with the existing token merging method. Furthermore, by leveraging the auxiliary spatial information, ToSA can generate more spatial coherent token merging results compared with previous work, as shown in Fig.~\ref{fig:teaser}, which further improves the question answering accuracy. The contributions of this work are summarized as follows:

\begin{itemize}
    \item We present \textbf{ToSA}, a training-free token merging method that conducted token merging based on semantic similarity and spatial awareness. We conduct extensive experiments on multiple images and video question answering benchmarks, demonstrating ToSA outperforms the previous semantic-based method ToMe with minimal additional computational cost, showing its potential for real-world applications.
    \item We present detailed qualitative results and merging comparison between previous work ToMe and ToSA, demonstrating ToSA can generate more spatial coherent merging results.
\end{itemize}
\section{Related Work}

\subsection{Efficient Vision Transformer}
To enhance the efficiency of ViT, most of the previous works adopt different ViT architectures that can downsample the number of visual tokens or reduce the computation of attention operations in the model. EfficientViT~\cite{efficientvit} introduced a cascaded group attention module and a parameter reallocation strategy to reduce the computational cost. DynamicViT~\cite{dynamicvit} and AdaViT~\cite{AViT} proposed additional learnable modules to reduce the number of visual tokens. SP-ViT~\cite{spvit} proposed spatial prior-enhanced attention to reduce the computational cost of self-attention blocks. However, these methods possess several limitations. Firstly, most of these methods lead to a dynamic number of visual tokens, which can not be applied to batch training and inference. Moreover, these models are optimized for image classification tasks, and their practicability for visual question answering task are not fully justified yet. Lastly, these methods' additional training parameters make them challenging to be directly adopted by recent VLMs~\cite{llava, onevision}, especially when most of the available model checkpoints are only trained on classification tasks. These reasons lead to the current VLMs still lean to leverage the original ViT architecture as visual encoder.

\subsection{Training-free Token Reduction}
In contrast to previous training-required methods, to enhance the practicability, several recent works do not require extra training and serve as a plug-and-play module to reduce the number of visual tokens and can enhance ViT and VLM efficiency. Work such as Evo-ViT~\cite{evovit} or EViT~\cite{EViT} utilized attention score to conduct token merging on less attentive tokens. ToMe~\cite{bolya2022tome} proposed bipartite soft matching and merged the visual tokens that share similar semantic visual features between each layer of ViT. These methods rely heavily on attention score or visual feature similarity as a criterion for token merging. However, as indicated by previous studies~\cite{vitlearn,ppt}, early layers of ViT features tend to demonstrate low-level information as well as similar attention scores across visual tokens within the same layer. This nature of ViT can potentially hinder the early-stage token merging process of the current methods, given the low-level similarity between two tokens does not guarantee they are from the same object. In addition to ViT-based token reduction methods, recent methods like Fast-V~\cite{fastv} and SparseVLM~\cite{sparsevlm} have introduced LLM-based visual token reduction conditioned on attention scores or text input. However, these approaches have several limitations. First, they offer weaker acceleration compared to ViT-based token reduction methods, as their token pruning occurs at a later stage in the VLM pipeline. Additionally, unlike ViT token reduction, LLM token reduction methods require storing all visual tokens across multiple conversation rounds, restricting their practicality in real-world applications. In this work, we focus on ViT-based token reduction methods for better practicality, aiming to provide more efficient acceleration while maintaining high performance on VLM's visual question answering tasks. Furthermore, instead of fully leveraging semantic affinity, we proposed to leverage spatial awareness as another criterion in the token merging process, aiming to improve the merging robustness of ViT in the earlier layers.

\begin{figure*}[t]
    \centering
    \includegraphics[width=0.98\linewidth]{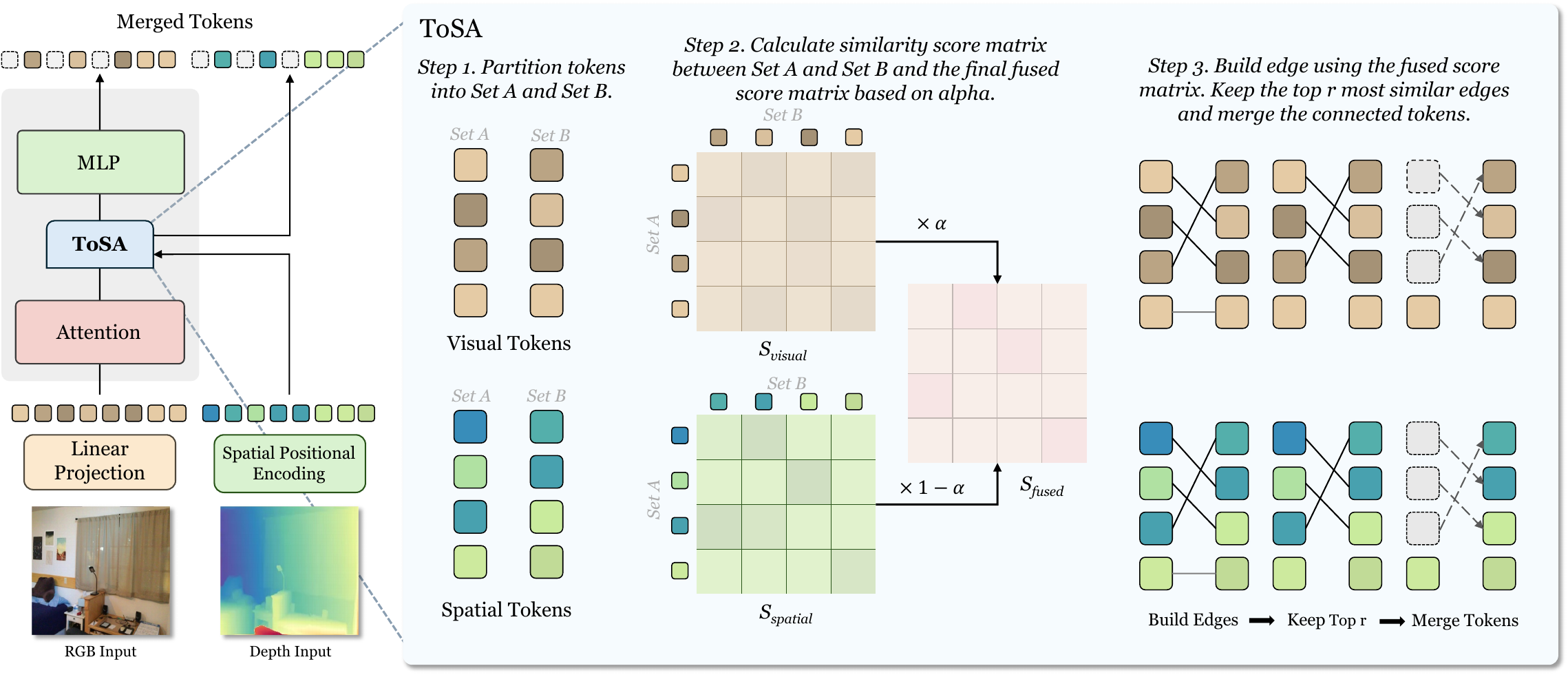}
    \caption{The overall framework of ToSA. ToSA block is inserted between attention and MLP across each encoder layer in ViT. ToSA block takes visual tokens and spatial tokens as input to conduct token merging. The merging process is based on the similarity of both visual tokens and spatial tokens.}
    \label{fig:tosa}
\end{figure*}

\subsection{Spatial Understanding}
Many recent works focus on advancing VLMs' ability to spatial understanding question answering tasks. SpatialVLM~\cite{chen2024spatialvlm} can conduct spatial understanding question answering from RGB input images after training on the collected large-scale spatial question answering data. SpatialRGPT~\cite{cheng2024spatialrgpt} and SpatialBot~\cite{spatialbot} incorporate extra depth modality as input to the visual encoder for spatial question answering. To evaluate the performance of state-of-the-art VLM on spatial understanding, SpatialBot further proposed SpatialBench, an image-based spatial understanding benchmark, featuring spatial understanding questions with multiple question categories including size comparison, counting, enumeration, spatial relationship...etc. On the other hand, VQAv2~\cite{vqa} includes object counting questions that can evaluate VLMs' capability in object counting, and GQA~\cite{gqa} focuses on more general image question answering tasks. OpenEQA~\cite{openeqa} is another benchmark that focuses on egocentric video-based 3D scene understanding, featuring RGB-D input with seven question categories, serving as a comprehensive benchmark for evaluating VLMs' spatial understanding capability in the 3D scene. In this work, we evaluated VLM's performance after incorporating our proposed ToSA on these spatial understanding benchmarks.
\section{Method}

\subsection{Preliminary}
Bipartite soft matching is an efficient way to merge the tokens within each ViT layer used by the previous token merging method~\cite{bolya2022tome}. It is applied on the visual tokens between the attention and MLP block as:
\begin{enumerate}
\setlength{\itemsep}{2pt}
\item Alternatively partition the tokens into two sets $\mathbb{A}$ and $\mathbb{B}$ of roughly equal size.
\item For each token in set $\mathbb{A}$, calculate the token similarity with each token in set $\mathbb{B}$ based on cosine similarity of the \textit{Key} features in the attention block. The similarity can be represented by a score matrix $\mathbf{S} \in \mathbb{R}^{|\mathbb{A}| \times |\mathbb{B}|}$, where $S_{ij}$ represents the cosine similarity between the $i^{th}$ token in $\mathbb{A}$ and the $j^{th}$ token in $\mathbb{B}$.
\item Merge the most similar $r$ pairs using a weighted average, and record the token size.
\item Concatenate the two sets $\mathbb{A}$ and $\mathbb{B}$ back together again.
\end{enumerate}

Once the tokens have been merged, they carry the features of more than one visual token. Therefore, the merged tokens will have less effect on softmax attention. We apply the proportional attention as:
\begin{equation} \label{eq:prop_attn}
    \bf A = \text{softmax}\left(\frac{{\bf Q}{\bf K}^\top}{\sqrt{d}} + \log s \right)
\end{equation}
where $\bf s$ is the number of patches the token represents after token merging.

\subsection{Spatial Token}
Given the visual token's feature is not reliable enough for merging in the early stage of ViT, we introduce spatial tokens, which are a series of tokens that are generated from the depth input. Spatial tokens contain rich spatial information and will go through the ToSA block along with the original visual tokens, serving as another criterion to calculate the score matrix for bipartite soft matching.

The way we generate spatial tokens is illustrated in Fig.~\ref{fig:tosa}. For each patch on the patchify depth image, we generate a triplet of index $(x,y,z)$, which stands for the index on the image. E.g. the most top left patch will have $x=0$ and $ y=0$, the $x$ and $y$ in the image ranging from the number of patches in $x$ and $y$ dimension. Here $x$ and $y$ range from $0$ to $26$, which is based on the resolution and patch size of our visual encoder. The third index $z$ is determined by the relative depth of the patch in the image, we average all the pixels' relative depth to obtain patch-wise relative depth. We divide the estimated depth value into 27 levels to match the index range in the $x$ and $y$ direction.

After the index assignment, each patch's spatial information will be represented by a triplet of index $(x,y,z)$, which we further encode with positional encoding following transformer~\cite{transformer}, each patch will turn into a spatial token that represents the patch spatial location, providing critical spatial prior for the ToSA token merging process.

\begin{table*}[t]
\centering
\caption{Category-level performance on SpatialBench.}
\resizebox{0.98\linewidth}{!}{
\begin{tabular}{lccccccc}
\toprule

Models & Position~($\uparrow$) &  Existence~($\uparrow$) & Counting~($\uparrow$) & Reaching~($\uparrow$) & Size~($\uparrow$)\\
\midrule
\textit{Base Model}\\
SpatialBot-3B & 58.8 & 80.0 & 86.7 & 53.3 & 25.0\\
\midrule
\textit{Token Merging - retain 10\% tokens}\\
ToMe & 52.9 & 55.0 & 58.7 & 58.3 & 25.0\\
ToSA & 52.9~(+0.0) & 65.0~(+10.0) & 72.7~(+14.0) & 63.3~(+5.0) & 21.7~(-3.3)\\

\bottomrule
\end{tabular}
}
\label{tab:spatialbench}
\end{table*}

\begin{table}
    \centering
    \caption{Performance comparison on the VQAv2 - counting.}
    \resizebox{0.68\linewidth}{!}{
    \begin{tabular}{l c c}
    \toprule
     Models & RCA & Acc.~(\%)\\
     \midrule
     \textit{Base Model} & \\
     LLaVA-OV-7B & 77.1 & 100\\
     \midrule
     \textit{Retain 50\% tokens} & \\
     ToMe & 53.1 & 68.9\\
     ToSA & 64.1 & 83.1\\
     \midrule
     \textit{Retain 10\% tokens} & \\
     ToMe & 54.1 & 70.2\\
     ToSA & 59.9 & 77.7\\
     \bottomrule
    \end{tabular}
    \label{tab:vqa}
    }
\end{table}
\begin{table}
    \centering
    \caption{Performance comparison on the GQA dataset.}
    \resizebox{0.7\linewidth}{!}{
    \begin{tabular}{l c}
    \toprule
     Models & EM@1\\
     \midrule
     BLIP-2~\cite{blip2} & 41.0 \\
     InstructBLIP~\cite{instructblip} & 49.2 \\
     Qwen-VL-Chat~\cite{qwenvl} & 57.5 \\
     LLaVA-1.5~\cite{llava1-5} & 62.0 \\
     LLaMA-VID~\cite{llamavid} & 62.3 \\
     VILA~\cite{vila} & 62.3 \\
     
     \midrule
     \textit{Base Model} & \\
     LLaVA-OV-7B & 66.0\\
     \midrule
     \textit{Retain 50\% tokens} & \\
     ToMe & 62.7\\
     ToSA & 63.2\\
     \midrule
     \textit{Retain 10\% tokens} & \\
     ToMe & 57.4\\
     ToSA & 57.8\\
     \bottomrule
    \end{tabular}
    \label{tab:gqa}
    }
\end{table}

\subsection{ToSA (Token Merging with Spatial Awareness)}
In the ViT forward pass, ToSA is inserted between the attention and MLP in the ViT layers to reduce the number of visual tokens. ToSA takes the visual tokens and spatial tokens as input, and calculates two separate score matrices $S_{visual}$ and $S_{spatial}$ based on the similarity of visual tokens and spatial tokens, respectively. Next, ToSA calculates the final fused score matrix based on the following equation, where $\alpha$ is a variable controlling the score matrix that focuses more on visual or spatial information.

\begin{equation}
    S_{fused} = \alpha S_{visual} + (1-\alpha)  S_{spatial}
\end{equation}

The fused score matrix $S_{fused}$ will be used by the bipartite soft matching, and the visual tokens and spatial tokens will be merged based on both visual and spatial information using the fused score matrix $S_{fused}$. Note that the visual tokens and spatial tokens will be merged in a corresponding style, i.e. if the $i^{th}$ and $j^{th}$ visual tokens are merged, the $i^{th}$ and $j^{th}$ spatial tokens will also be merged accordingly.

The parameter $\alpha$ is a weighting factor that controls whether ToSA relies more on visual feature similarity or spatial affinity. In the early layers, given the visual features are less reliable, we use a smaller $\alpha$ to encourage ToSA to conduct token merging based on spatial affinity. In the deeper layers, we use a larger $\alpha$ so that ToSA can utilize the more semantic features in the deep layers of ViT. This merging strategy seamlessly takes advantage of both semantic-based merging and the auxiliary spatial prior provided by the RGB-D input. In the default setting, we use a linear increase schedule for $\alpha$, which gradually increases when passing through the layers of ViT. The value of $\alpha$ at layer $i^{th}$ can be denoted as:

\begin{equation}
    \alpha_{i} = {i}/{L}
\label{eq:alpha}
\end{equation}

\noindent with $\alpha_{i}$ represents the $\alpha$ value in the $i^{th}$ layer and $L$ is the total number of layers in ViT. We also conducted different schedules of $\alpha$ values and their effect on the performance of spatial understanding, see Table.~\ref{tab:schedule} for more results.
\section{Experiments}
\label{sec:exp}

\subsection{Implementation}
We use SpatialBot-Phi2-3B~\cite{spatialbot} in our experiments on SpatialBench. On other benchmarks, we use LLaVA-OneVision-7B~\cite{onevision} as our base model. Both VLMs use siglip-so400m-patch14-384~\cite{siglip} as the visual encoder, which has 27 ViT layers, and $L$ in Eq.~\ref{eq:alpha} is set to 27 accordingly. SpatialBot-Phi2-3B is trained on the SpatialQA~\cite{spatialbot}, and LLaVA-OneVision-7B is trained on single, multi-image, and video data collected from a wide range of publicly available instruction following dataset. For embodied question answering benchmarks, we uniformly sampled 12 images from the 3D scans as input to our VLM. The depth image used to generate the spatial tokens is predicted by depth-anything-v2~\cite{depthanythingv2}. All experiments are conducted on a V100 GPU.

\subsection{Benchmarks}
There are several publicly available spatial understanding question answering benchmarks, including 
SpatialRGPT-Bench~\cite{cheng2024spatialrgpt}, and SpatialBench~\cite{spatialbot}. SpatialRGPT-Bench focuses on dense region spatial question answering, which requires extra region boxes or masks as input to the ViT and can not be seamlessly adopted by most of the current VLMs. Therefore, we used SpatialBench, which features 5 different question types related to object position, existence, counting, reaching, and size. Besides SpatialBench, we also evaluated ToSA on counting questions collected from VQAv2~\cite{vqa} and GQA~\cite{gqa}, which focuses on more general visual reasoning questions. Besides image-level question answering, we further evaluated ToSA on embodied question answering tasks, which involve video-level question answering in 3D environments. We conducted our experiments on the OpenEQA~\cite{openeqa} dataset, which addresses embodied question answering across seven categories, covering aspects such as object recognition, spatial reasoning, object localization, attribute recognition...etc. Additionally, OpenEQA incorporates an LLM scorer that evaluates the quality of predictions by comparing them with ground truth, offering assessments that better align with human judgment.

\subsection{Performance}
Most of the efficient ViT methods~\cite{efficientvit,dynamicvit,AViT} required re-training the ViT for different tasks beyond image classification, and there are very limited VLM integrations that can enable us to conduct spatial question answering evaluation using these models. For this reason, we mainly compared our proposed ToSA with another recent training-free, off-the-shelf ViT token merging method ToMe~\cite{bolya2022tome}. ToMe has been used by many VLM~\cite{auroracap,longvlm} for efficient training and inference, which demonstrates its successful integration for image and video understanding tasks.\\

\noindent{\textbf{SpatialBench.}} In Table.~\ref{tab:spatialbench}, we listed the category-level performance of base model SpatialBot-3B~\cite{spatialbot} and its performance after applying different token merging methods. We demonstrate that ToSA can achieve comparable performance with ToMe in most of the tasks, and largely outperforms ToMe on task that requires more fine-grained spatial understanding such as object existence and counting. The question consists of multi-choice questions and counting, where the former is evaluated based on accuracy. The latter is calculated with Relative Counting Accuracy, which is calculated as \( 1 - \frac{|x - y|}{y} \), where \( x \) is the predicted count and \( y \) is the ground truth.\\

\noindent{\textbf{VQAv2-Counting.}} In Table~\ref{tab:vqa}, we present the performance on VQAv2's counting questions. We adopt SpatialBench's Relative Counting Accuracy~(RCA), and also report the accuracy percentage with respect to the base model.\\

\noindent{\textbf{GQA.}} In Table.~\ref{tab:gqa}, we listed the performance of different VLMs and the comparison between ToSA and ToMe. Despite GQA focusing more on general question answering, instead of heavily related to spatial question answering, ToSA achieves better performance compared with ToMe across different visual token usage.\\

\noindent{\textbf{OpenEQA.}} We show the experiment results on OpenEQA in Table.~\ref{tab:openeqa}, including the performance of proprietary VLMs, and several open-source VLM's performance with some of them~\cite{videochatgpt,videollama,videollama2} integrate extra learnable modules that can conduct token reduction for video understanding task.

\begin{table}
    \centering
    \caption{Performance comparison on the OpenEQA dataset.}
    \resizebox{0.98\linewidth}{!}{
    \begin{tabular}{l c c}
    \toprule
     Models & LLM-Match\\
     \midrule
     \textit{Proprietary VLMs} & \\
     Claude-3 Opus & 36.3 \\
     Gemini 1.0 Pro Vision & 44.9 \\
     Claude-3.5 Sonnet & 48.7 \\
     GPT4-V (15 frames) & 54.6 \\
     GPT4-V (50 frames) & 55.3 \\
     \midrule
     \textit{Open-source VLM} & \\
     Video-LLaMA~\cite{videollama} & 20.0 \\
     LLaMA-2 w/ Concept Graph~\cite{openeqa} & 28.7 \\
     AuroraCap~\cite{auroracap} & 28.9\\
     Video-ChatGPT~\cite{videochatgpt} & 32.1 \\
     LLaMA-2 w/ Sparse Voxel Map~\cite{openeqa} & 34.3 \\
     LLaMA-2 w/ LLaVA-1.5 caption~\cite{openeqa} & 36.8 \\
     Chat-UniVi~\cite{jin2024chat} & 42.3\\
     Video-LLaMA2~\cite{videollama2} & 49.2 \\
     \midrule
     \textit{Base Model} & \\
     LLaVA-OV-7B & 56.2\\
     \midrule
     \textit{Retain 10\% tokens} & \\
     ToMe & 48.3\\
     ToSA & 49.5\\
     \bottomrule
    \end{tabular}
    \label{tab:openeqa}
    }
\end{table}
\begin{table}[t]
    \centering
    \begin{minipage}{0.4\linewidth}
        \small
        \centering
        \caption{Ablation on token merging schedule.}
        \begin{tabular}{l c}
        \toprule
        Schedule & Acc\\
        \midrule
        uniform & 56.7 \\
        decrease & 52.8 \\
        increase & \textbf{59.9} \\
        \bottomrule
        \end{tabular}
        \label{tab:schedule}
    \end{minipage}%
    \hfill
    \begin{minipage}{0.59\linewidth}
        \small
        \centering
        \caption{Ablation on inference speed.}
        \begin{tabular}{ccc}
        \toprule
        Methods & Used Token & im/s\\
        \midrule
        SigLIP & 100\% & 18.2\\
         \midrule
         ToMe   &  50\% & 23.8 \\
         ToSA   &  50\% & 23.7 \\
         \midrule
         ToMe   &  10\% & 33.7 \\
         ToSA   &  10\% & 33.5 \\
        \bottomrule
        \end{tabular}
        \label{tab:speed}
    \end{minipage}
\end{table}

\begin{figure*}[t]
    \centering
    \includegraphics[width=0.98\linewidth]{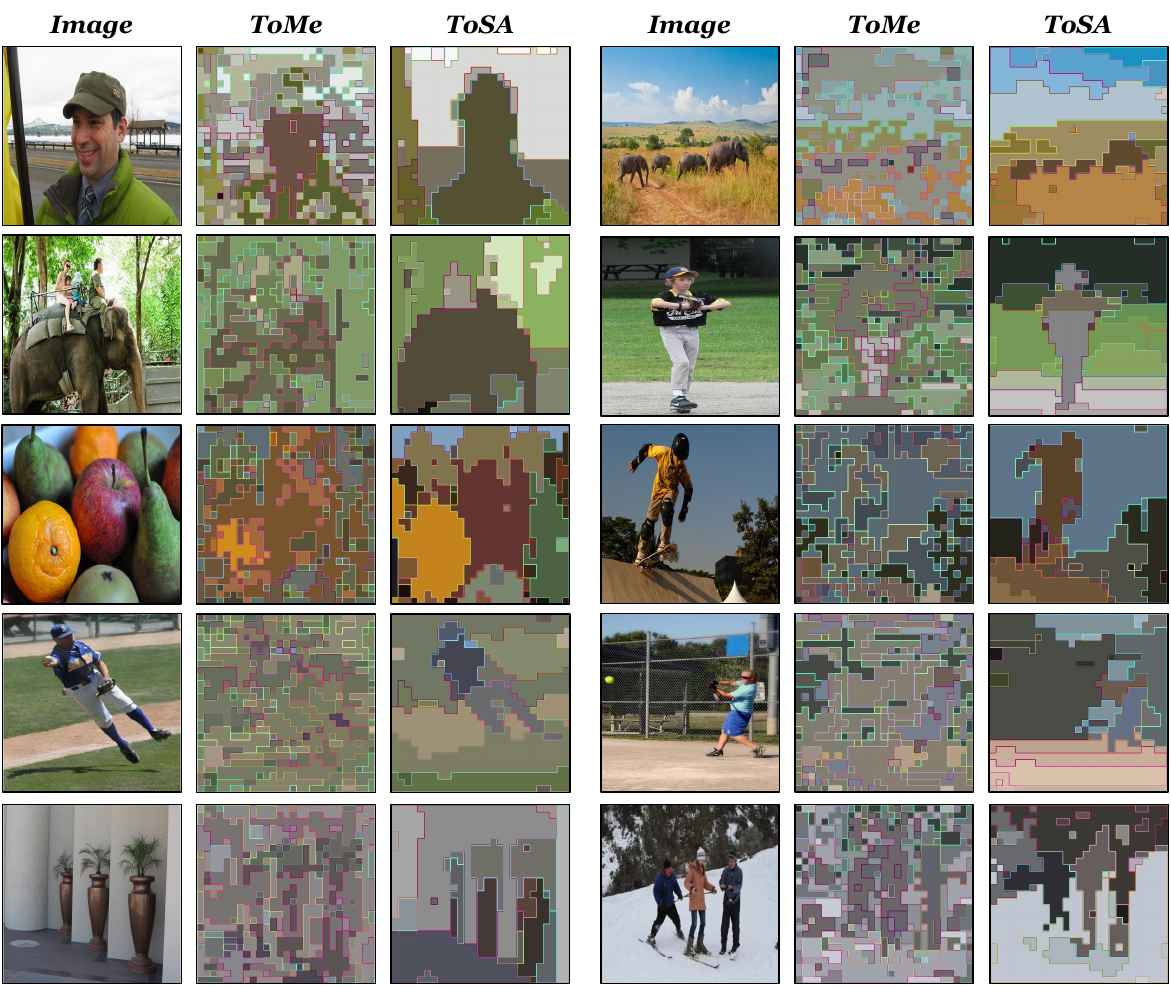}
    \caption{Merging results comparison between ToMe and ToSA. We only keep 16 visual tokens~(2\%) from both methods for better visual comparison. ToSA demonstrate more spatially coherent merging results, which leads to higher performance across various visual question answering benchmarks.}
    \label{fig:qual}
\end{figure*}

\subsection{Ablation Studies}
\noindent{\textbf{Effect of Alpha Schedule.}} To demonstrate that the earlier layer's visual features are suboptimal for token merging, we conduct experiments based on different $\alpha$ schedules in Table~\ref{tab:schedule}. We tested three different alpha schedules, including a uniform $\alpha$ value across all layers, where $\alpha$ is set to as constant across all layers. We set $\alpha$ to $0.5$ for our experiment. We also evaluate the decrease schedule where the $\alpha$ starts with $1$ and gradually reduces to $0$ in the deeper layers of ViT. As shown in Table~\ref{tab:schedule}, using the increase schedule for $\alpha$ resulted in the highest accuracy on VQAv2-counting, and outperforms the other two schedules by a large margin. This showcases our assumption that it is beneficial to leverage spatial awareness in earlier layers for token merging.\\
\noindent{\textbf{Inference Speed.}} We compared the inference speed of ToMe and ToSA in Table~\ref{tab:speed}. Both methods are tested on the V100 GPU in our experiments. Both ToMe and ToSA can achieve a large boost in throughput compared with the original ViT. Compared with ToMe, ToSA incorporates extra spatial tokens during the inference, but ToSA's degradation of throughput is minimal, with less than 0.6\% degradation under 50\% used token and 10\% used token settings. Note that the throughput is different from ToMe's original reported number because our SigLIP uses a higher-resolution image input, which results in more visual tokens.

\subsection{Qualitative Results}
We showcase some qualitative results in Fig.~\ref{fig:qual}, which exhibits the difference in merging results between semantic-only ToMe and ToSA's semantic and spatial-aware merging. Because ToSA leverages spatial awareness during the merging process, the qualitative results demonstrate ToSA maintains a more coherent token merging results, with respect to the spatial structure of the image. Here, the same token is denoted by the same color and inner edges. Note that we only keep 16 visual tokens (2\%) from both methods for better visual comparison between the two token merging methods.


\section{Conclusion}
In this work, we proposed ToSA, a training-free token merging method that conducts token merging based on spatial prior. Experimental results show that ToSA can achieve better performance compared with previous token merging methods on multiple VQA benchmarks including SpatialBench, VQAv2, GQA, and OpenEQA. Furthermore, despite the improvements, ToSA introduces minimal additional inference cost, with less than 0.6\% of runtime degradation compared with previous work, demonstrating its potential for real-world applications.

\section{Limitation}
Although ToSA demonstrates advanced performance compared with the existing method, ToSA requires auxiliary depth image as input during the merging process. This makes ToSA more applicable to scenarios such as robotic or autonomous driving, where RGB-D input data is available.

\clearpage

\bibliographystyle{IEEEtran}
\bibliography{IEEEabrv,ref}

\end{document}